\documentclass{article}

    \PassOptionsToPackage{numbers, compress}{natbib}

    \usepackage[preprint]{neurips_2025}

\usepackage[utf8]{inputenc} 
\usepackage[T1]{fontenc}    
\usepackage{url}            
\usepackage{booktabs}       
\usepackage{amsfonts}       
\usepackage{nicefrac}       
\usepackage{microtype}      
\usepackage{xcolor}         

\usepackage{multirow}
\usepackage{graphicx}
\setcitestyle{numbers}
\usepackage{amsmath}
\usepackage{soul}
\usepackage{tcolorbox}
\usepackage[hidelinks]{hyperref}

\title{IQBench: How ``Smart'' Are Vision-Language Models? A Study with Human IQ Tests}

\author{Tan-Hanh Pham$^{1,\dag, \ddag}$, Phu-Vinh Nguyen$^{2,\dag}$, Dang The Hung$^3$,\\\textbf{Bui Trong Duong}$^4$, \textbf{Vu Nguyen Thanh}$^5$, \textbf{Chris Ngo}$^6$, \textbf{Tri Quang Truong}$^5$, \textbf{Truong-Son Hy}$^7$\\
\hphantom{text}\\ 
$^1$Harvard Medical School, USA,
$^2$Uppsala University, Sweden,
$^3$University of London, UK,\\
$^4$Vietnam Military Medical University,
$^5$University of Technical Education Ho Chi Minh City, Vietnam,\\
$^6$Knovel Engineering Lab, Singapore, $^7$University of Alabama at Birmingham, USA\\
\hphantom{text}\\ 
$^\dag$Equal contribution; $^\ddag$Corresponding author
}

\begin{document}

\maketitle

\begin{abstract}
Although large Vision-Language Models (VLMs) have demonstrated remarkable performance in a wide range of multimodal tasks, their true reasoning capabilities on human IQ tests remain underexplored. To advance research on the fluid intelligence of VLMs, we introduce \textbf{IQBench}, a new benchmark designed to evaluate VLMs on standardized visual IQ tests. \textbf{We focus on evaluating the reasoning capabilities of VLMs}, which we argue are more important than the accuracy of the final prediction. \textbf{Our benchmark is visually centric, minimizing the dependence on unnecessary textual content}, thus encouraging models to derive answers primarily from image-based information rather than learned textual knowledge. To this end, we manually collected and annotated 500 visual IQ questions to \textbf{ prevent unintentional data leakage during training.} Unlike prior work that focuses primarily on the accuracy of the final answer, we evaluate the reasoning ability of the models by assessing their explanations and the patterns used to solve each problem, along with the accuracy of the final prediction and human evaluation. Our experiments show that there are substantial performance disparities between tasks, with models such as \texttt{o4-mini}, \texttt{gemini-2.5-flash}, and \texttt{claude-3.7-sonnet} achieving the highest average accuracies of 0.615, 0.578, and 0.548, respectively. However, all models struggle with 3D spatial and anagram reasoning tasks, highlighting significant limitations in current VLMs' general reasoning abilities. In terms of reasoning scores, \texttt{o4-mini}, \texttt{gemini-2.5-flash}, and \texttt{claude-3.7-sonnet} achieved top averages of 0.696, 0.586, and 0.516, respectively. These results highlight inconsistencies between the reasoning processes of the models and their final answers, emphasizing the importance of evaluating the accuracy of the reasoning in addition to the final predictions. Code and data are publicly available at \href{https://anonymous.4open.science/r/IQBench_anonymous-3515/}{\textcolor{teal}{here}}.
\end{abstract}

\section{Introduction}

The rapid evolution of artificial intelligence (AI) has been driven by the development of Large Language Models (LLMs), Multimodal Models, and Vision-Language Models (VLMs), which have significantly improved natural language understanding and cross-modal reasoning capabilities \cite{radford2021learning, pham2025silvar}. Early LLMs, such as GPT-2 and GPT-3 \cite{brown2020language}, demonstrated strong performance in tasks such as text generation and question answering. Building on this progress, multimodal models such as CLIP \cite{radford2021learning} were developed to integrate visual and textual data, allowing tasks such as image captioning and zero-shot classification. More recently, open-source VLMs, including BLIP-2, PaLI, Flamingo, and Qwen-VL \cite{li2023blip, chen2023pali, alayrac2022flamingo, bai2023qwen}, have pushed the boundaries of AI by allowing systems to process and reason over both visual and textual inputs with high accuracy. In parallel, industry efforts have introduced advanced multimodal models such as OpenAI's chatbot series (GPT-4o, o1, o3, o4-mini), Google's Gemini family, Anthropic's Claude, and X's Grok, which further enhance multimodal understanding and generation.

These advancements have fueled interest in Artificial General Intelligence (AGI), defined as the ability to perform any intellectual task a human can undertake \cite{goertzel2007artificial}, which is based on fluid intelligence, the ability to solve novel problems through abstract reasoning \cite{cattell1987intelligence, chollet2019measure}. Evaluating whether these models exhibit AGI-like intelligence requires robust metrics that encompass accuracy, reasoning transparency, and generalizability, supported by advanced data centers and evaluation frameworks \cite{liang2022holistic, srivastava2022beyond}.

Numerous benchmarks have been developed to assess the VLM capabilities in diverse tasks. Benchmarks such as Visual Question Answering (VQA) \cite{antol2015vqa}, GQA \cite{hudson2019gqa}, ScienceQA \cite{lu2022learn}, and Visual Commonsense Reasoning (VCR) \cite{zellers2019recognition} evaluate a model’s ability to interpret visual scenes, while benchmarks such as MMMU \cite{yue2024mmmu}, MathVista \cite{lu2024mathvista}, and ChartQA \cite{masryetal2022chartqa} focus on mathematical and analytical reasoning from visual input. Other efforts, such as Image2Struct \cite{roberts2024image2struct}, challenge models with low-vision tasks like extracting structure from images or rendering code based on visual content. However, these benchmarks are predominantly based on the accuracy of the final answer, which often does not reveal the reasoning process behind the predictions \cite{rajpurkar2018know}. Studies have shown that high accuracy can stem from the use of biases in the dataset or shortcut learning rather than genuine reasoning \cite{agrawal2018don, geirhos2020shortcut}. With recent advancements in reasoning-focused models such as OpenAI's o1, o3, o4-mini, GPT-4o \cite{zhong2024evaluation, hurst2024gpt}, Anthropic’s Claude 3.5 Sonnet \cite{anthropic2024claude}, DeepSeek-R1 \cite{guo2025deepseek}, Google’s Gemini 2.5 Flash \cite{google2024gemini}, and Grok-3 from xAI \cite{xai2024grok3}, evaluating whether these models exhibit such models requires robust metrics that go beyond surface-level performance, including accuracy, reasoning transparency and generalizability.

Despite the increasing number of VLM benchmarks, significant limitations persist, particularly in evaluating fluid intelligence. Many benchmarks are nearing saturation, with state-of-the-art models achieving near-human performance in tasks such as VQA \cite{lu2022learn} and greater accuracy 85\% in math and coding challenges \cite{zhong2024evaluation}. This trend, along with the rise of reasoning models, suggests that existing benchmarks are increasingly ineffective in challenging VLMs or exposing their reasoning deficiencies. Although benchmarks such as MMLU \cite{wang2024mmlu} and ARC \cite{chollet2019measure} assess reasoning, they are limited to textual tasks and lack the multimodal complexity of visual IQ tests. Even newer multimodal benchmarks like MathVista \cite{lu2024mathvista} and MMMU \cite{yue2024mmmu} do not specifically target fluid intelligence or reasoning interpretability, and recent multimodal models have already exceeded 80\% accuracy in MMMU, suggesting that its effectiveness is becoming saturated. In particular, some models achieve strong results, such as Sphinx-X-MoE scores 43.6\% on MMMU without images, outperforming its LLM backbone (17.9\%) \cite{chen2024we}. Such findings risk overestimating true multimodal ability and highlight weaknesses in current benchmarks.

To address these challenges, we introduce \textbf{IQBench}, a novel vision-centric benchmark designed to evaluate the fluid intelligence of VLMs using standardized visual IQ tests. IQBench comprises curated questions that consist of IQ tests, where each question is annotated with the correct answer and a detailed reasoning pattern, facilitating in-depth analysis of the behavior of the model. Unlike previous work, we propose a dual-metric evaluation framework comprising a \textit{\textbf{reasoning score}}, derived using an LLM-as-judge methodology to assess the accuracy and coherence of a model's explanation, and a \textit{\textbf{accuracy score}} for the final predictions. Complemented by human evaluations, this framework offers a comprehensive assessment of VLM performance. Our evaluation of the leading VLMs on IQBench provides critical insight into their reasoning capabilities, laying the groundwork for developing more transparent and cognitively robust multimodal systems.

Our contributions are summarized as follows:
\begin{itemize}
    \item \textbf{IQBench benchmark for fluid intelligence}: We introduce IQBench, a novel benchmark specifically designed to evaluate the fluid intelligence of VLMs using curated visual IQ test questions.
    \item \textbf{Vision-Centric and manually curated dataset}: To avoid unintentional data leakage during model training, we curated and annotated 500 visual IQ questions, including reasoning patterns and correct answers across a wide range of topics.
    \item \textbf{Reasoning evaluation framework}: We evaluated the prediction of the model using both an accuracy score and a reasoning score to gain deeper insight into their reasoning capabilities.
    \item \textbf{Comprehensive analysis of state-of-the-art VLMs}: Through extensive experiments, we evaluated the leading VLMs on IQBench and uncover their strengths and weaknesses in various reasoning tasks.
\end{itemize}



\section{IQBench}

We introduce \textbf{IQBench}, a novel benchmark designed to evaluate the fluid intelligence of VLMs through standardized visual IQ tests. IQBench comprises 500 human-curated questions that encompass a comprehensive range of IQ test domains, including pattern recognition, analogical reasoning, visual arithmetic, spatial understanding, abstract/concrete reasoning, number/figure series reasoning, anagrams, and verbal reasoning with syllogisms. Each question is annotated with the correct answer and a detailed reasoning pattern, allowing a granular analysis of the behavior of the model. Unlike existing benchmarks that focus primarily on the accuracy of the final response, IQBench emphasizes both the accuracy of predictions and the interpretability of the reasoning process, addressing critical gaps in evaluating VLM capabilities \cite{lei2024generalized, bi2025verify, cai2025mm}.

\subsection{Data Collection and Generation}

The IQBench dataset was constructed through a rigorous process of manual data collection and generation to ensure diversity and originality. We collect images from various sources, including online repositories, textbooks, and educational materials, which were then edited or used as inspiration to generate new questions and reasoning patterns with the topics indicated in Table~\ref{tab:iqbench_stats}. As shown in the table, the dataset contains 500 samples, divided equally into 10 reasoning topics, with 50 questions per topic. All images are saved in PNG format to keep them clear and consistent.

\begin{table}[h]
    \centering
    \caption{Dataset statistics for IQBench.}
    \begin{tabular}{ll}
        \toprule
        \textbf{Metric} & \textbf{Value} \\
        \midrule
        Total samples & 500\\
        Number of topics & 10 \\
        Total sample of each topic & 50\\
        Image type & PNG \\
        \midrule
        \multicolumn{2}{l}{Question and answer format:}\\
        Multiple choice &  110  \\
        Open questions  &  390 \\
        Average question length (words) & 27 \\
        Average pattern length (words) & 48 \\
        \bottomrule
    \end{tabular}
    \label{tab:iqbench_stats}
\end{table}

The questions are presented in two formats: multiple choice (110 questions) and open-ended (390 questions), with a focus on open-ended questions to better test the model’s reasoning ability. On average, each question is 27 words long, and each reasoning pattern is about 48 words. This helps provide a strong base for evaluating how well models can understand and explain their answers in different types of visual IQ problem.

\begin{figure}[h]
    \centering
    \includegraphics[width=0.985\linewidth]{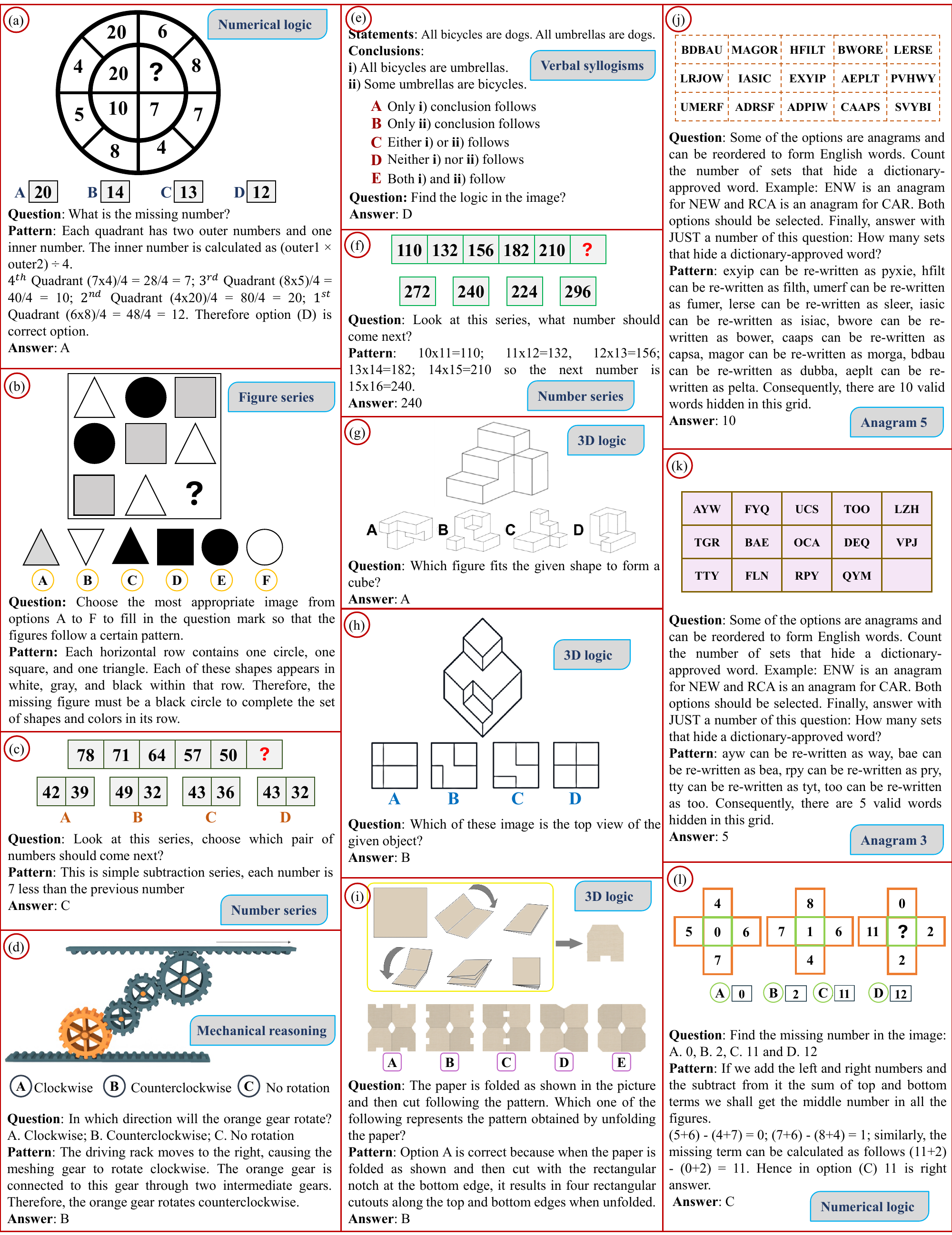}
    \caption{Representative IQ test covering logic, pattern recognition, and spatial reasoning for VLM evaluation. A sample consists of an input image, a question, an answer, and a possible reasoning pattern.}
    \label{fig:dataset}
\end{figure}

To build IQBench, we followed a structured process that ensured both legal compliance and originality of the content. First, we collected raw visual materials from verified sources, ensuring that all content was in compliance with copyright regulations. When images were of low quality or their copyright status was unclear, we generated new visuals to maintain both clarity and legality. Next, we created original questions inspired by standard visual IQ tests. Each question was carefully designed to assess different reasoning skills and was manually annotated with the correct answer along with a detailed reasoning pattern.

\subsection{Data Quality Control}

To ensure the reliability and fairness of IQBench, we focused on maintaining human-generated high-quality content. This helps reduce data leakage and the likelihood that models have encountered similar examples during pre-training, which could lead to inflated performance. To address the redundancy of images in many existing benchmarks, we designed IQBench to be vision-centric, minimizing the possibility of models answering questions based solely on language knowledge rather than visual understanding.

We also applied lexical overlap analysis to detect and remove potential duplicate samples, ensuring the uniqueness of each question. In addition, we conducted thorough manual reviews to standardize the format of all questions and answers, promoting consistency and clarity. An overview of the data is presented in Fig.~\ref{fig:dataset}, where each sample contains an input image, a question, an answer, and a possible associated reasoning pattern. For example, some samples do not contain patterns such as in Fig.~\ref{fig:dataset}g and Fig.~\ref{fig:dataset}h.

\subsection{Evaluation Methods}
Unlike previous benchmarks, which primarily assess the accuracy of the final answer, our evaluation emphasizes both the reasoning process and the final answer. This dual focus is particularly effective for evaluating models on multiple-choice question answering tasks. For instance, a model might hallucinate during the reasoning process but still select a plausible answer (e.g., ``A'' from options A, B, C, D), resulting in an uninformative accuracy 25\% without any insight into its reasoning. To address these limitations, IQBench introduces a dual-metric evaluation framework that comprehensively assesses VLM performance. This includes:
\begin{itemize}
    \item \textbf{Accuracy Score}: This metric evaluates the correctness of the model's final prediction via an exact match. It is applicable to both multiple-choice and open-ended responses, enabling standardized comparisons with existing benchmarks while capturing a model’s ability to arrive at the correct answer.
    \item \textbf{Reasoning Score}: This score assesses the coherence, correctness and alignment of the model’s explanation with the expected reasoning path. We adopt an LLM-as-judge strategy, where pre-trained LLMs compare the model’s explanation with annotated reasoning patterns for each question. This metric quantifies interpretability and sheds light on the cognitive path leading to the answer, an aspect often overlooked in traditional evaluations.
\end{itemize}
For both metrics, we use \texttt{gpt-4o-mini} as the judge model, assigning a score of 1 for correct responses and 0 for incorrect ones. In addition to these automated metrics, we include \textbf{human evaluations} to assess the reasoning of VLM. This human-in-the-loop component provides a qualitative benchmark to complement automated metrics (LLM-as-judge). We suppose that this multi-faceted evaluation approach offers a holistic understanding of VLM capabilities, balancing performance accuracy with reasoning transparency.

\textbf{Comparison with existing VLM benchmarks}, IQBench distinguishes itself by emphasizing fluid intelligence and the interpretability of reasoning across a diverse spectrum of cognitive tasks. Unlike benchmarks such as MMMU \cite{yue2024mmmu} and MathVista \cite{lu2024mathvista}, which focus on domain-specific reasoning (e.g., academic or mathematical), IQBench is designed to assess general intelligence using a variety of abstract, symbolic, and logic-based problems inspired by human IQ tests. Although MM-IQ \cite{cai2025mm}, ARC \cite{chollet2019measure}, and Verify \cite{bi2025verify} contribute valuable insights into model accuracy, they focus primarily on the correctness of the answers without systematically evaluating the traceability of the reasoning or the breadth of cognitive skills. In contrast, IQBench aims to measure not only what the model answers but also how it reasons, providing a more comprehensive evaluation of VLM intelligence.

\section{Experiment and Result}
To evaluate the fluid intelligence of VLMs in IQBench, we performed experiments using a zero-shot testing framework. We tested seven state-of-the-art VLMs: \texttt{gemini-2.5-flash}, \texttt{gemini-2.0-flash}, \texttt{claude-3.7-sonnet}, \texttt{claude-3.5-sonnet}, \texttt{gpt-4o}, \texttt{o4-mini}, and \texttt{gpt-o3}. These models were selected for their advanced multimodal capabilities and recent advancements in reasoning-focused architectures. 

\subsection{Accuracy Evaluation}

\begin{table*}[ht]
\centering
\caption{Evaluation accuracy of models on IQBench tasks. Task abbreviations: MDRT (Mechanical Deductive Reasoning Test), DRTF (Deductive Reasoning Test with Figures), 3D SPRT (3D Spatial Deductive Reasoning Test), VRTS (Verbal Reasoning Test with Syllogisms), IVRT (Inductive Verbal Reasoning Test), Num. (Numerical), FS (Figure Series Test), NS (Number Series Test), Ana5 (Anagrams 5 Test), Ana3 (Anagrams 3 Test), Avg. (Average score).}
\label{tab:iqbench_accuracy}
\resizebox{\textwidth}{!}{
\begin{tabular}{lccccccccccc}
\toprule
\textbf{Model} &
\textbf{MDRT} &
\textbf{DRTF} &
\textbf{3D SPRT} &
\textbf{VRTS} &
\textbf{IVRT} &
\textbf{Num.} &
\textbf{FS} &
\textbf{NS} &
\textbf{Ana5} &
\textbf{Ana3} & \textbf{Avg.}\\
\midrule
\texttt{gemini-2.5-flash} & 0.60 & 0.78 & 0.18 & 0.66 & 0.78 & 0.74 & 0.60 & 0.88 & 0.14 & 0.42 &  0.578\\
\texttt{gemini-2.0-flash} & 0.58 & 0.56 & 0.22 & 0.70 & 0.74 & 0.52 & 0.44 & 0.84 & 0.16 & 0.14 &  0.490\\
\texttt{claude-3.7-sonnet} & 0.64 & 0.90 & 0.40 & 0.72 & 0.68 & 0.46 & 0.66 & 0.82 & 0.04 & 0.16 &  0.548\\
\texttt{claude-3.5-sonnet} & 0.62 & 0.68 & 0.20 & 0.74 & 0.74 & 0.32 & 0.42 & 0.76 & 0.12 & 0.10 &  0.470\\
\texttt{gpt-4o} & 0.56 & 0.42 & 0.20 & 0.80 & 0.74 & 0.36 & 0.26 & 0.66 & 0.06 & 0.02 &  0.408\\
\texttt{o4-mini} & 0.72 & 0.86 & 0.34 & 0.66 & 0.76 & 0.82 & 0.60 & 0.94 & 0.02 & 0.14 &  0.615\\
\texttt{gpt-o3} & 0.70 & 0.88 & -- & -- & -- & -- & -- & -- & 0.12 & -- & -- \\
\bottomrule
\end{tabular}
}
\end{table*}

\begin{figure}[h]
    \centering
    \includegraphics[width=0.8\linewidth]{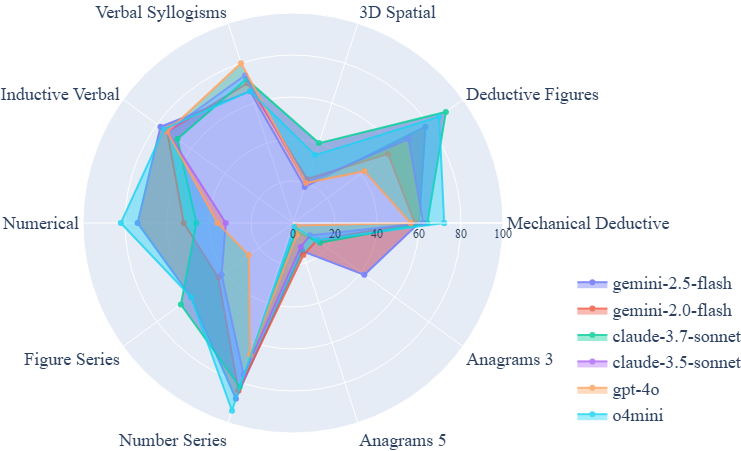}
    \caption{Accuracy evaluation of the advanced multimodal models on IQBench.}
    \label{fig:accuracy_result}
\end{figure}

The experimental results are reported in Table~\ref{tab:iqbench_accuracy}, showing the accuracy of various models in IQBench tasks. The highest overall performance was achieved by \texttt{o4-mini} with an average accuracy of 0.615, followed by \texttt{gemini-2.5-flash} (0.578) and \texttt{claude-3.7-sonnet} (0.548). These models performed particularly well on tasks such as Number Series (NS), Deductive Reasoning with Figures (DRTF), and Verbal Reasoning with Syllogisms (VRTS), with several scores exceeding 0.80. For example, \texttt{o4-mini} achieved 0.94 in NS and 0.86 in DRTF.

Despite this, all models struggled on tasks that required advanced spatial reasoning and linguistic manipulation. Specifically, the 3D Spatial Deductive Reasoning Test (3D SPRT) and the Anagram tasks (Ana5 and Ana3) revealed consistent weaknesses. For example, \texttt{gemini-2.5-flash} scored only 0.18 on 3D SPRT and 0.14 on Ana5, while \texttt{gpt-4o} performed the worst overall with 0.408 average accuracy and only 0.02 on Ana3. Intuitively, the general coverage and task-level performance distribution of the models are illustrated in Fig.~\ref{fig:accuracy_result}.

\textbf{An important observation here is that although these multimodal models excel in text-based reasoning, they struggle with the anagram task}, which requires identifying meaningful English words from scrambled letters, indicating limitations in their fine-grained linguistic manipulation capabilities. These findings suggest that even advanced VLMs face substantial challenges in tasks that test fluid intelligence components such as mental rotation, abstract symbol manipulation, and semantic reconfiguration, highlighting key areas for future model improvement.

\subsection{Reasoning Evaluation}
Table~\ref{tab:iqbench_reasoning} shows the reasoning scores, where \texttt{o4-mini} achieved the highest average score (0.696), followed by \texttt{gemini-2.5-flash} (0.586) and \texttt{claude-3.7-sonnet} (0.516). In particular, \texttt{o4-mini} excel in tasks that require structured reasoning, such as the Mechanical Deductive Reasoning Test (MDRT: 0.92) and the Figure Series Test (FS: 0.90), which demonstrate robust explanatory coherence.

A key trend observed across models is that reasoning scores tend to be slightly higher than accuracy scores for several models. For example, \texttt{gpt-4o} shows a notable gap between accuracy (0.408) and reasoning (0.466), and similarly, \texttt{claude-3.7-sonnet} has a reasoning score (0.516) nearly matching its accuracy (0.548), despite variability in task difficulty. This suggests that many models can produce logically sound or plausible reasoning chains even when the final selected answer is incorrect, potentially due to misinterpretation of visual content or confusion in multi-choice mapping.

\begin{figure}[h]
    \centering
    \includegraphics[width=0.8\linewidth]{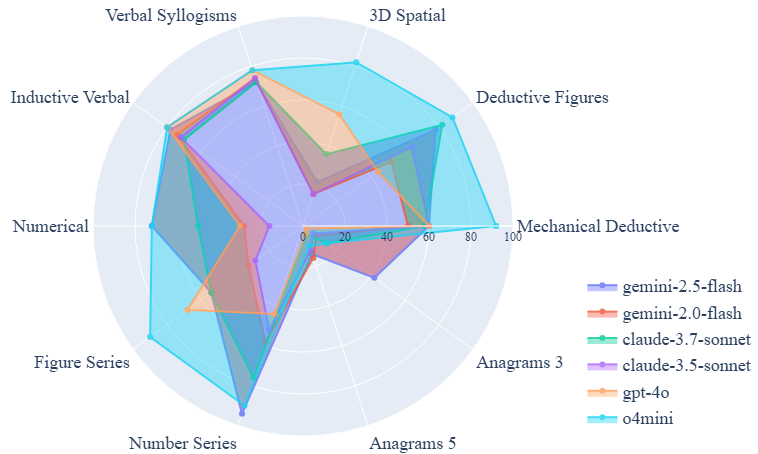}
    \caption{Reasoning evaluation of the advanced multimodal models on IQBench.}
    \label{fig:Reasoning_score}
\end{figure}

In contrast, models such as \texttt{gemini-2.0-flash} and \texttt{claude-3.5-sonnet} show both low accuracy and reasoning scores, indicating more fundamental limitations in both understanding and explaining visual reasoning tasks. Meanwhile, \texttt{o4-mini} stands out as the only model with strong and balanced performance in both dimensions, indicating a well-aligned multimodal architecture capable of accurate predictions and coherent justifications. Intuitively, the general coverage and the reasoning performance distribution of the models are illustrated in Fig.~\ref{fig:Reasoning_score}.

\begin{table*}[ht]
\centering
\caption{Reasoning Evaluation of Models on IQBench Tasks.}
\label{tab:iqbench_reasoning}
\resizebox{\textwidth}{!}{
\begin{tabular}{lccccccccccc}
\toprule
\textbf{Model} &
\textbf{MDRT} &
\textbf{DRTF} &
\textbf{3D SPRT} &
\textbf{VRTS} &
\textbf{IVRT} &
\textbf{Num.} &
\textbf{FS} &
\textbf{NS} &
\textbf{Ana5} &
\textbf{Ana3} &
\textbf{Avg.} \\
\midrule
\texttt{gemini-2.5-flash} & 0.60 & 0.78 & 0.22 & 0.72 & 0.78 & 0.72 & 0.54 & 0.94 & 0.14 & 0.42 &  0.586\\
\texttt{gemini-2.0-flash} & 0.50 & 0.52 & 0.16 & 0.74 & 0.74 & 0.28 & 0.32 & 0.58 & 0.16 & 0.08 &  0.408\\
\texttt{claude-3.7-sonnet} & 0.58 & 0.82 & 0.36 & 0.72 & 0.70 & 0.50 & 0.54 & 0.76 & 0.04 & 0.14 &  0.516\\
\texttt{claude-3.5-sonnet} & 0.60 & 0.64 & 0.16 & 0.74 & 0.72 & 0.16 & 0.28 & 0.52 & 0.12 & 0.06 &  0.400\\
\texttt{gpt-4o} & 0.60 & 0.44 & 0.56 & 0.78 & 0.80 & 0.30 & 0.68 & 0.44 & 0.04 & 0.02 &  0.466\\
\texttt{o4-mini} & 0.92 & 0.88 & 0.82 & 0.78 & 0.80 & 0.72 & 0.90 & 0.90 & 0.10 & 0.14 &  0.696\\
\texttt{gpt-o3} & 0.70 & 0.88 & -- & -- & -- & -- & -- & -- & 0.12 & -- & -- \\
\bottomrule
\end{tabular}
}
\end{table*}

\begin{tcolorbox}
[colback=black!3!white,colframe=teal!50!white,boxrule=0.3mm,title=, left=0mm,right=0mm]
\textbf{Observation}: (1) Models perform well on tasks like DRTF and NS, which involve pattern recognition and numerical reasoning, but struggle with 3D SPRT and Anagrams, indicating gaps in spatial and vision-centric linguistic reasoning. (2) Reasoning scores are often slightly higher than accuracy scores, showing that models can sometimes explain their incorrect answers in a logically coherent way, highlighting a disconnect between reasoning quality and decision correctness. (3) \texttt{o4-mini} and \texttt{gemini-2.5-flash} demonstrate superior performance in both metrics, likely due to greater visual-textual integration and alignment in their architectures.
\end{tcolorbox}

\subsection{Human Evaluation and Failure Analysis}
Among the seven models evaluated in our study, \texttt{o4-mini} consistently achieved the highest performance in multiple reasoning tasks. Due to its strong results, we selected \texttt{o4-mini} for a focused human evaluation to better understand the quality and interpretability of its reasoning process. To ensure a manageable and representative assessment, we randomly sampled 100 predictions from the model for manual review. Three human experts independently evaluated the performance of the model and the results are summarized in Table~\ref{tab:human_evaluation}, where a reasoning score of 1 is assigned for correct reasoning and 0 for incorrect reasoning.

\begin{table*}[ht]
\centering
\caption{Comparison of reasoning scores assigned by human experts and LLM-as-judge (\texttt{gpt-4o-mini}) for \texttt{o4-mini} predictions across IQBench tasks.}
\label{tab:human_evaluation}
\resizebox{\textwidth}{!}{
\begin{tabular}{lccccccccccc}
\toprule
\textbf{Experts} &
\textbf{MDRT} &
\textbf{DRTF} &
\textbf{3D SPRT} &
\textbf{VRTS} &
\textbf{IVRT} &
\textbf{Num.} &
\textbf{FS} &
\textbf{NS} &
\textbf{Ana5} &
\textbf{Ana3} &
\textbf{Avg.} \\
\midrule
\multicolumn{10}{l}{\textit{LLM-as-Judge}} \\
\texttt{gpt-4o-mini} & 0.92 & 0.88 & 0.82 & 0.78 & 0.80 & 0.72 & 0.90 & 0.90 & 0.10 & 0.14 &  0.696\\
\midrule
\multicolumn{10}{l}{\textit{Human evaluation}}\\
Expert 1 & 0.90 & 0.80 & 0.20 & 1.00 & 0.90 & 0.89 & 0.90 & 1.00 & 0.90 & 0.80 & 0.83 \\
Expert 2 & 0.70 & 0.90 & 0.30 & 0.70 & 0.90 & 0.80 & 0.90 & 1.00 & 0.00 & 0.10 & 0.63 \\
Expert 3 & 0.80 & 0.80 & 0.20 & 0.70 & 0.60 & 0.80 & 0.78 & 0.90 & 0.10 & 0.20 & 0.59 \\
\cmidrule{2-12}
\texttt{Average} & 0.80 & 0.83 & 0.23 & 0.80 & 0.80 & 0.83 & 0.86 & 0.97 & 0.33 & 0.37 & 0.68 \\

\bottomrule
\end{tabular}
}
\end{table*}

Table~\ref{tab:human_evaluation} presents the reasoning scores of \texttt{o4-mini} as assessed by human experts and \texttt{gpt-4o-mini} as the judge. The average human-assigned reasoning score for all tasks is 0.68, while the LLM judge score is 0.696, demonstrating a close alignment between the two evaluation methods. In most categories, LLM as a judge scores fall within the range of human evaluations, indicating a high level of agreement in assessing the quality of the reasoning. In particular, the consistency in trend between tasks (e.g., higher scores in FS and NS, and lower scores in Ana5 and Ana3) further supports the validity of the LLM-as-judge method. These results suggest that the automated evaluation framework is a reliable proxy for human judgment in assessing the interpretability and correctness of VLM-generated reasoning, making it a scalable and practical alternative for large-scale benchmarking.

\begin{figure}[ht]
    \centering
    \includegraphics[width=1\linewidth]{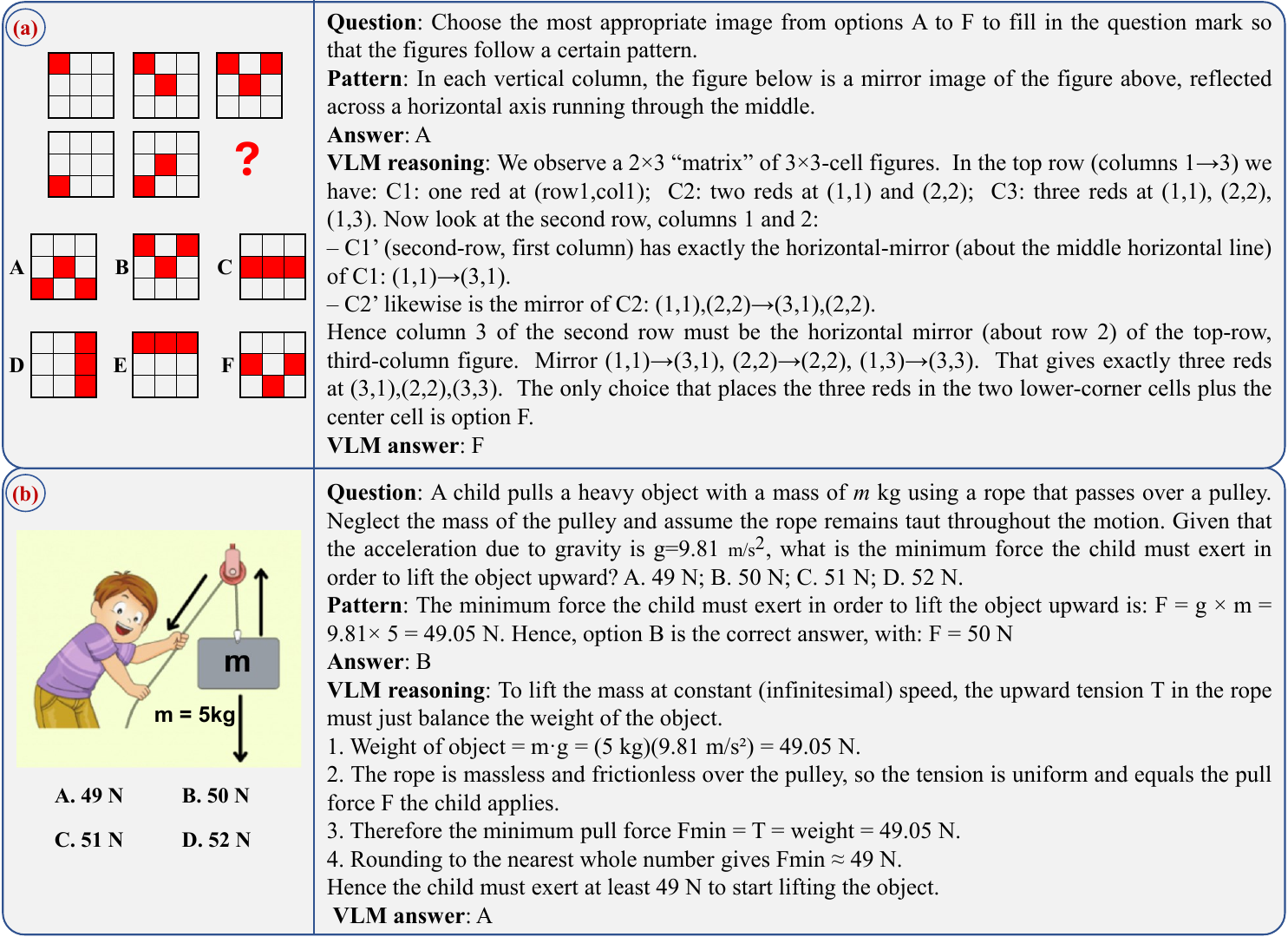}
    \caption{Examples of VLM reasoning errors compared to human annotations. (a) Correct reasoning with incorrect prediction in a pattern recognition task. (b) Partially correct reasoning with incorrect prediction in a physics problem.}
    \label{fig:result}
\end{figure}

In human evaluation, we also point out some incorrect reasoning of the model as shown in Fig. \ref{fig:result}. For the pattern recognition task (Fig. \ref{fig:result}a), the VLM correctly identifies the horizontal mirror reflection pattern in a matrix $2 \times 3$ but selects option F instead of A, probably due to misinterpreting the visual options. For the physics problem (Fig. \ref{fig:result}b), the VLM accurately calculates the force ($49.05 \, \text{N}$) but rounds incorrectly to 49 N (option A) instead of 50 N (option B), reflecting the failure to choose the minimum force. These examples highlight the model's limitations in aligning its reasoning with visual options and in integrating multiple input modalities to arrive at the correct answer.

\section{Conclusion}
We present \textbf{IQBench}, a novel benchmark designed to evaluate the fluid intelligence of Vision-Language Models through vision-centric IQ questions. Unlike existing benchmarks that emphasize answer accuracy alone, IQBench promotes both \textit{answer correctness} and \textit{reasoning interpretability} by introducing a two-fold evaluation framework. Our results show that even the most advanced VLMs, such as o4-mini, Claude 3.5 Sonnet, and Gemini 1.5 Flash, struggle with key reasoning categories, particularly 3D spatial understanding and anagram tasks. Moreover, the quality of reasoning often does not align with the correctness of the final answers, highlighting the current limitations in the cognitive depth and generalization of VLMs. To advance multimodal model development, we believe that IQBench will foster research toward more transparent, robust, and cognitively capable VLMs, ultimately bringing us closer to AGI systems with genuine problem-solving abilities.

\section*{Limitation}
\label{sec:limitation}
Despite the benefit of our benchmark in evaluating Vision-Language Models and their reasoning ability, our work still contains some limitations, including the use of the LLM-as-a-judge pipeline to assess reasoning ability and the limited number of samples in our dataset.


\bibliographystyle{unsrtnat}
\bibliography{ref}

\newpage
\appendix
\label{sec:appendix}
\section{Prompts for General VLMs}

\begin{verbatim}
Given the image, answer the following question: {question}
Your answer must include your reasoning and strictly follow this format:
<reason>
Your thinking to find the final answer of the problem
</reason>
<answer>
Your final answer
- For multiple choice, answer with a letter (A, B, C, etc.).
- For numerical or computed answers, answer with a number.
</answer>

**IMPORTANT**
- your answer must include <reason> and <answer> sections
- your answer must start with <reason> and end with </answer>
- <reason> section provide your thinking to answer the question
\end{verbatim}

\section{Prompts for Reasoning VLMs}

\begin{verbatim}
Given the image, answer the following question: {question}
Your answer must strictly follow this format:
<answer>
Your final answer
- For multiple choice, answer with a letter (A, B, C, etc.).
- For numerical or computed answers, answer with a number.
</answer>

**IMPORTANT**
- your answer must include <answer> section
- your answer must start with <answer> and end with </answer>
\end{verbatim}

\section{LLM-as-a-judge Prompt}

\begin{verbatim}
# Given the following information:

## Question (You will not able to see the image as you should only compare 
the groud truth thinking and VLM's reasoning)
{question}

## Ground Truth Reasoning (GT Reasoning)
{pattern}

## Ground Truth Final Answer (GT Answer)
{answer}

## VLM's Reasoning (VLM Reasoning)
{think}

## VLM's Final Answer (VLM Answer)
{bot_answer}

## Task
Your task is to analyze whether the VLM's reasoning is logically sound and 
consistent with the ground truth reasoning, and whether it leads to the 
correct final answer. Base your judgment on reasoning accuracy, logical 
consistency, and whether the intermediate steps support the final conclusion.

Respond strictly in the following format:
<reason>
Compare the VLM's reasoning to the ground truth reasoning. Is the logical 
structure similar?
Are the key steps present? Does the reasoning correctly support the final 
answer? Mention any discrepancies or alignments.
</reason>
<evidence>
If the VLM's reasoning is flawed or deviates from the ground truth, provide 
specific parts of the VLM reasoning that are incorrect, missing, or misleading. 
If correct, explain why the reasoning is logically valid.
</evidence>
<answer>
Return 1 if the VLM's reasoning is correct and aligns with the ground truth, 
otherwise return 0.
</answer>

**IMPORTANT**
- your answer must include <answer> section
- contents inside the <answer> section must be just 1 or 0
- your answer must start with <reason> and end with </answer>
\end{verbatim}


\end{document}